\documentclass{article}

 \usepackage[final]{nips_2017}

\usepackage[utf8]{inputenc} % allow utf-8 input
\usepackage[T1]{fontenc}    % use 8-bit T1 fonts
\usepackage{hyperref}       % hyperlinks
\usepackage{url}            % simple URL typesetting
\usepackage{booktabs}       % professional-quality tables
\usepackage{amsfonts}       % blackboard math symbols
\usepackage{nicefrac}       % compact symbols for 1/2, etc.
\usepackage{microtype}      % microtypography
\usepackage{subcaption}      % microtypography

\usepackage{graphicx} % more modern
\usepackage{amsmath}

\newcommand\E{\mathbb{E}}
\newcommand\Var{\text{Var}}
\newcommand\Cov{\text{Cov}}

\newcommand\nper{{n_{\text{per}}}}
\DeclareMathOperator*{\plim}{plim}

\title{Learning causal effects from many randomized experiments using regularized instrumental variables}

\author{
Alexander Peysakhovich\\
Facebook AI Research
\And
Dean Eckles \\
Massachusetts Institute of Technology\\
}

\begin{document}

\maketitle

\begin{abstract}
Scientific and business practices are increasingly resulting in large collections of randomized experiments. Analyzed together, these collections can tell us things that individual experiments in the collection cannot. We study how to learn causal relationships between variables from the kinds of collections faced by modern data scientists: the number of experiments is large, many experiments have very small effects, and the analyst lacks metadata (e.g., descriptions of the interventions). Here we use experimental groups as instrumental variables (IV) and show that a standard method (two-stage least squares) is biased even when the number of experiments is infinite. We show how a sparsity-inducing $l_0$ regularization can --- in a reversal of the standard bias--variance tradeoff in regularization --- reduce bias (and thus error) of interventional predictions. Because we are interested in interventional loss minimization we also propose a modified cross-validation procedure (IVCV) to feasibly select the regularization parameter. We show, using a trick from Monte Carlo sampling, that IVCV can be done using summary statistics instead of raw data. This makes our full procedure simple to use in many real-world applications.
\end{abstract}

\section{Introduction}
Randomized experiments (i.e. A/B tests, randomized controlled trials) are a popular practice in medicine, business, and public policy \citep{banerjee2012poor, kohavi2013online}. When decision-makers employ experimentation they have a far greater chance of learning true causal relationships and making good decisions than via observation alone \citep{lalonde1986evaluating, meyer2015two, hemkens2016agreement}. However, a single experiment is often insufficient to learn about the causal mechanisms linking multiple variables --- which in turn can be important for theory building and/or decision-making.

Consider the situation of a internet service for watching videos. The firm is interested in how watching different types of videos (e.g., funny vs. serious, short vs. long) affects user behaviors (e.g. by increasing time spent on the site, inducing subscriptions, etc.). This will inform decisions about content recommendation or content acquisition. Even though the firm can measure all relevant variables, learning a model on observational data will likely be misleading; for example, existing content recommendation systems and heterogeneous user dispositions will produce strong correlations between exposure to many video types and time spent or subscription, but it is not true that the magnitude of this correlation is the response that the company can expect if they \textit{intervene} and change the promotion or availability of videos. Thus, we are interested not just in prediction but prediction under intervention \citep{bottou2013counterfactual, bottou2014machine, pearl2009causality}.

The standard solution here is to run a randomized experiment exposing some users to more of some type of video. However, a single A/B test will likely change many things in the complex system. It is hard to change the number of views of funny videos without affecting the number of views of serious videos or short videos. This problem is sometimes called `fat hand' interventions because we touch multiple causal variables at once. This means the firm likely cannot learn a vector of causal effects (one for each video type) in such a simple manner. Thus, the company would need to use multiple A/B tests together (e.g., in a factorial design). 

However, because routine product experimentation is common in internet companies \citep{bakshy2014www,varian2016intelligent,kohavi2013online}, this firm has likely already run many A/B tests, including on the video recommendation algorithm. The method proposed in this paper can either be applied to a new set of experiments run explicitly to learn a causal effect vector \citep[as in, e.g.,][]{eckles2016estimating}, or can be applied to repurpose already run tests by treating them as random perturbations injected into the system and using that randomness in a smart way. 

Our contributions arise from adapting the econometric method of instrumental variables \citep[IV;][]{wright1928tariff, reiersol1945confluence, angrist1996identification} to this setting. It is well known that a standard IV estimator --- two-stage least squares (TSLS) --- is biased in finite samples \citep{stock2012survey, angrist2008mostly}. For our case, it also has asymptotic bias. We show that this bias depends on the distribution of the treatment effects in the set of experiments under consideration.

Our main technical contribution is to introduce a multivariate $l_0$ regularization into the first stage of the TSLS procedure and show that it can reduce the bias and MSE of estimated causal effects. Because in finite samples this regularization procedure reduces bias but adds variance, we introduce a method to select this regularization parameter which we call \emph{instrumental variables cross-validation} (IVCV). In an empirical evaluation that combines simulation and data from hundreds of real randomized experiments, we show that the $l_0$ regularization with IVCV outperforms TSLS and a Bayesian random effects model.

Finally, we show how to perform this estimation in a computationally and practically efficient way. Like standard TSLS, our regularization and cross-validation procedures only require summary statistics at the level of experimental groups. This is advantageous when using raw data is computationally or practically burdensome, e.g., in the case of internet companies. This means the computational and data storage complexities of the method are actually quite low. In addition, standard A/B testing platforms \citep{bakshy2014www, xu2015infrastructure} should already compute and store all the required statistics, so the method here can be thought of as an ``upcycling'' of existing statistics.

\section{Confounding and the Basic IV Model}
Suppose we have some (potentially vector valued) random variable $X$ and a scalar valued outcome variable $Y$. We want to ask: what happens to $Y$ if I change some component of $X$ by one unit, holding the rest constant? Formally, we study a linear structural (i.e. data generating) equation pair
$$X = U\psi + \epsilon_{X} \text{ and } Y = X\beta + U\gamma + \epsilon_{Y}$$
where $U, \epsilon_X$, and $\epsilon_Y$ are independent random variables with mean 0, without loss of generality. Note that in A/B testing we are often interested in relatively small changes to the system, and thus we can just think about locally linear approximations to the true function. We can also consider basis expansions.
We refer to $X$ as the causal variables (in our motivating example this would be a vector of time spent on each video type), $Y$ as the outcome variables (here overall user satisfaction), $U$ as the unobserved confounders, $\epsilon$ as noise, and $\beta$ as the causal effects. 

In general, we are interested in estimating the causal effect $\beta$ because we are interested in intervention, e.g., one which will change our data-generating model to $X = U\psi + \epsilon_{X} + a.$ 

In the presence of unobserved confounders, $\beta$ is not identified and trying to learn causal relationships using predictive models naively can lead us astray \citep{bottou2013counterfactual, bottou2014machine, shalit2016bounding, pearl2009causality}. Suppose that we have observational data of the form $(X, Y)$ with $U$ completely unobserved. If we use this data to estimate the causal effect $\beta$ we can, due to the influence of the unobserved confounder, get an estimate that is (even in infinite samples) larger, smaller or even the opposite sign of the true causal effect $\beta$ (we describe this more fully in the Supplemental Material). Thus, the best predictor of $Y$ given $X$ may not be lead to a good estimate of what would happen to $Y$ if we \textit{intervened}.

We now discuss instrumental variable (IV) estimator as a method for learning the causal effects. Suppose that we have some variable $Z$ that has two properties. First, $Z$ is not caused by anything in the $(X, U, Y)$ system; that is, $Z$ is as good as randomly assigned. Second, $Z$ affects $Y$ only via $X$. This latter assumption is known as an exclusion restriction or complete mediation assumption. Formally, this modifies the structural equation (see the Supplemental Material for the DAG representation) for $X$ to be
$$
X = Z \mu + U \psi + \epsilon_{X}
$$

The standard IV estimator for $\beta$ is two-stage least squares (TSLS) and works off the principle that the variance in $X$ can be broken down into two components. The first component is confounded with the true causal effect (i.e. comes from $U$). The second component, on the other hand, is independent of $U$. Thus, if we could regress $Y$ only on the random component, we could recover the causal effect $\beta$. Knowing $Z$ allows us to do exactly this (i.e. by using only the variation in $X$ caused by $Z$ not $U$).

TSLS can be thought of as follows: in the first stage we regress $X$ on $Z$. We then replace $X$ by the predicted values from the regression. In the second stage, we regress $Y$ on these fitted values.\footnote{
We make an additional assumption: in order to estimate the effect of each variable $X_{j}$ on $Y$ with the other $X$'s held constant it must be the case that $Z$ is such that it causes independent variation in all dimensions of $X$. This means that we must, at least, have as many instruments as the dimension of $\beta$ for TSLS to work.
%An interesting and fruitful direction for future work is what to do when some subspace of $X$ is well spanned by our instruments but some some subspace is not.
} It is straightforward to show that as $n$ approaches infinity this estimator converges to the true causal effect $\beta$ \citep[Theorem 5.1]{wooldridge2010econometric}.

\section{IV with Test Groups without Metadata}
In our setting of interest, randomly assigned groups from a large collection of experiments are the instruments. That is, the IV is a categorical variable indicating which of $K$ test groups a unit (e.g., user) was assigned to in one of many experiments. For simplicity of notation, we assume that each treatment group $g \in \{1, ..., K \}$ has exactly $n_g = \nper$ units assigned to it at random.

\subsection{Computational Properties}
The way to represent the first stage regression of the TSLS is to use the \emph{one-hot representation} (or dummy-variable encoding) of the group which each unit is assigned to, such that $Z_i$ is a $K$-dimensional vector of 0s and a single 1 indicating the randomly assigned group.

In this setup the TSLS estimator has a very convenient form. The first stage regression of $X$ on $Z$ simply yields estimates that are group level means of $X$ in each group. This means that if each group has the same number of units (e.g., users) and the same error variance, the second stage has a convenient form as well: we can recover $\beta$ by simply regressing group level averages of $X$ on $Y$ \citep[section 4.1.3]{angrist2008mostly}.

Thus, to estimate causal effects from large meta-analyses practitioners do not need to retain or compute with the raw data (which can span millions or billions of rows in the context of A/B testing at a medium or large internet company), but rather can retain and compute with sample means of $X$ and $Y$ in each A/B test group (this is now just thousands of rows of data). These are quantities that are recorded already in the most automated A/B testing systems \citep{bakshy2014www, xu2015infrastructure}. Working with summary statistics simplifies computation enormously and allows us to reuse existing data.

\subsection{Asymptotic Bias in the Grouped IV Estimator}
There are now multiple ways to think about the asymptotic properties of this ``groups as IVs'' estimator. Either we increase the size of each experiment ($\nper \rightarrow \infty$) or we get more experiments ($K \rightarrow \infty$). The former is the standard asymptotic sequence, but for meta-analysis of a growing collection of experiments, the latter is the more natural asymptotic series, so we fix $\nper$ but we raise $K$.

We fix ideas with the case where $X, Y, Z, U$ are scalar. We denote the group level means of our variables with bars (e.g., $\bar{X}$ to be the random variable that is the group-level means of $X$). Recall that our TSLS is, in the group case, a regression of $\bar{Y}$ on $\bar{X}$.

Decompose the causal variable group level average into $\bar{X} = \bar{Z} + \bar{U} \psi + \epsilon_{\bar{X}},$ where $\bar{Z} \equiv Z \mu = \E[X | Z]$ is the true first stage of the IV model (i.e. what we are trying to learn in the first stage of the TSLS).  In the case of experiments as instruments this term has a nice interpretation --- it is the true average value of the causal variables when assigned to that experimental group.

While we are not considering asymptotic series where $\nper$ goes to infinity, $\nper$ will generally also be large enough that so that we can use the normality of sample means guaranteed by the central limit theorem. Thus, $\bar{U}$ and $\bar{\epsilon}_X$ are normal with mean $0$ and variance proportional to $\frac{1}{n_{per}}.$

With finite $\nper$ we can show that, even as $K \to \infty$, TSLS will be biased \citep[cf.][]{bekker1994alternative, angrist1995split}. Suppose for intuition that $\bar{Z}$ has mean $0$ and finite variance $\sigma^2_{\bar{Z}}$ this bias has the closed form (see Supplemental Materials for a derivation of the general form):
$$
\plim_{K \to \infty} \hat{\beta}_\text{TSLS} = \beta + \frac{\gamma \psi \frac{ \sigma^2_{U}}{\nper}}{\psi^{2} \frac{\sigma^2_U}{\nper}  + \frac{\sigma^2_{\epsilon_X}}{\nper} + \sigma^2_{\bar{Z}}}.
$$
To understand where this bias comes from, think about the case where $\bar{Z}$ is always $0$. The instrument does nothing, however the group-level averages still include group-level confounding noise; that is, for finite $\nper$, $\bar{U}$ has positive variance. Thus, we simply recover the original observational estimate that we have already discussed as including omitted variable bias. When $Z$ is not degenerate, $\bar{X}$ and $\bar{Y}$ include variation from both $\bar{U}$ and $\bar{Z}$. As $\nper$ increases the influence of $\bar{U}$ decreases and so $\hat{\beta}_{\text{TSLS}}$ is consistent for $\beta.$\footnote{While in many cases, where variation induced by instrumental variables is large, this bias can be safely ignored, in the case of online A/B testing this is likely not the case. Since much of online experimentation involves hill climbing and small improvements (on the order of a few percent or less) that add up, the TSLS estimator can be quite biased in practice (more on this below).}

\section{Bias-Reducing Regularization}
We now introduce a regularization procedure that can decrease bias in the TSLS estimator. We show that, in this setting a $l_0$-regularized first stage is computationally feasible and can help reduce this bias under some conditions on the distribution of the latent treatment effects.

\subsection{Intuition via a Mixture Model}
There are many types of A/B tests conducted --- some are micro-optimizations at the margin and some are larger explorations of the action space. Consider the stylized case with two types of tests calling the smaller variance type `weak' tests while the larger variance ones are `strong' test, where the type gives the distribution from which its treatment effects are drawn; that is, $\bar{Z}$ is drawn from a two-component mixture model, with probability $p_{\text{weak}}$, we have that $\bar{Z}$ has variance $\sigma^2_{\text{weak}}$ and with probability $(1-p_{\text{weak}})$ it has variance $\sigma^2_{\text{strong}}$. 

Notice that if we ran TSLS using only groups whose $\bar{Z}$ is drawn from component $j \in \{\text{weak}, \text{strong} \}$, then our estimator converges to
$$
\plim_{K \to \infty} \hat{\beta}_{\text{TSLS}, j}  = \beta + \gamma \frac{ \psi \frac{\sigma^2_{U}}{\nper}}{\psi^{2} \frac{\sigma^2_U}{\nper}  + \frac{\sigma^2_{\epsilon_{X}}}{\nper} + \sigma^2_{j}}
$$
Because $\sigma^2_\text{strong} > \sigma^2_\text{weak}$ we will have that $\hat{\beta}_{\text{TSLS, strong}}$ is a less biased estimator than $\hat{\beta}_{\text{TSLS, weak}}.$ If we don't know which test is of which type and simply run a TSLS on the full data set, we will get some estimator that will be a weighted combination of these two quantities. Thus, with sufficient number of groups, we can actually improve our causal estimate by using less data (i.e. only the strong tests). Of course when the number of tests $K$ is finite we face a bias--variance tradeoff.

Within this discrete mixture model, we are limited to how much we can reduce bias (since $\plim_{K \to \infty} \hat{\beta}_\text{TSLS, strong} \neq \beta$). However suppose that the treatment effects are drawn from a distribution which is an infinite mixture of normals that has full support on normals of all variances, such as a $t$ distribution, then we can asymptotically (in the large $K$ sense) reduce the bias below any $\epsilon$ by using only observations which come from components with arbitrarily large variances. We now introduce a regularization procedure to do this.

\subsection{Formalizing First Stage Regularization}
Consider a data set $(\bar{X}_g, \bar{Y}_g)$ of vectors of group-level averages.
Let
$$
p(x) = \Pr(|\bar{U}_g + \bar{\epsilon}_{x,g}| > |x|)
$$
be the $p$-value for a group-level observation $x$ under a `no intervention' null with $Z = 0$. These are straightforward to compute from the observational (i.e., within control condition) variance (or covariance matrix) of $X$.
For a given threshold $q \in (0,1]$, let 
$$
\bar{X}^q_g \equiv
\begin{cases}
   \bar{X}_g& \text{if } p(\bar{X}_g) < q\\
    0              & \text{otherwise}.
\end{cases}$$
We then define the regularized IV estimator as
$$\hat{\beta}_q = (\bar{X}^{q'} {\bar{X}}^q)^{-1} (\bar{X}^{q'} \bar{Y}).$$

Thus, this procedure is equivalent to an $l_0$ regularization in the first stage of the TSLS regression. In particular, when $\bar{U}_g + \bar{\epsilon}_{x,g}$ has a normal distribution, as in the present case, then this is equivalent to $l_0$-regularized least squares.

Recall that in the binary mixture example above, this regularization would preferentially retain groups that come from the higher variance (strong) component. This extends to infinite mixtures, such as the $t$, where this procedure will preferentially set $\bar{X}_g$ to zero for groups where $\bar{Z}_g$ is drawn from a lower variance component.

So far we have focused on scalar $X$. This procedure naturally extends to multidimensional settings. Compute $p(\bar{X}_g)$ and simultaneously threshold all dimensions of the experimental group $g$; that is, if this probability is above a threshold $q$ we set the whole vector $\bar{X}_g$ to $0.$ This is thus a group-$l_0$ regularizer.\footnote{
We note that this group-$l_0$ regularization is inefficient if treatment effects are such that each A/B test only only moves a single dimension of $X$ (i.e. 'skinny hand' interventions). In our evaluation we see that it works in real world applications, however, it is an interesting question for future research to learn its limitations. See the Supplemental Material for additional simulations and discussion.}

%Recall that the bias term of interest is given by $\gamma \frac{\Cov(\bar{X}^q, \bar{U})}{\Var(\bar{X}^q)}.$
%As we move the threshold $q$ to a lower $p$-value (corresponding to a larger $|\bar{X}_g|$ value) we `hollow out' the middle of the distribution of $\bar{X}_g$ and increase the conditional variance of $\Var(\bar{X}_q \mid p(\bar{X}_g^q) > q)$. However, the numerator also changes, indeed as we condition on higher levels of $|\bar{X}|$, we also get higher levels of $|\bar{U}|$ and thus the covariance between them increases. We have a race condition: if the numerator increases slower than the denominator increases, then as we take the threshold $q$ all the way to $0$, we also lower the bias to $0$. We have seen in the section above how this effect can operate when the distribution is a mixture of normals of different variances.\footnote{A common test used in practice for whether an TSLS estimate is credible is the size of the $F$ or Cragg-Donald statistic in the first stage regression \citep{angrist2008mostly, stock2005testing, stock2012survey}. Often a threshold value is used. However, the discussion above shows that even when these traditional tests may say that a TSLS is too biased (e.g., if $\bar{Z}$ comes from a mixture with lots of mass at 0), when $K$ is large there may still be hope.}

\section{Causal Cross-Validation}
We now turn to an important practical question: because there is a bias--variance tradeoff how should one set the regularization parameter when $K$ is finite to optimize for prediction under intervention? 

First, let us suppose that we have access to the raw data where a row is a $(X_i, Z_i, Y_i)$ which is a unit $i$'s, $X$, $Y$ and treatment assignment $Z.$ We propose a procedure to set our hyperparameter $q$. We describe $2$-fold version as it conveys the full intuition, but extension to $k$-folds is straightforward. 

\textbf{Instrumental variables cross-validation algorithm} (IVCV):
\begin{enumerate}
\item Split \textit{each treatment} in the data set into $2$ folds, call these new data sets $\lbrace (X^1_i, Y^1_i, Z^1_i) \rbrace$ and $\lbrace  (X^2_i, Y^2_i, Z^2_i) \rbrace$.
\item Compute treatment level averages $\lbrace (\bar{X}^1_g, \bar{Y}^1_g) \rbrace$ and $\lbrace (\bar{X}^2_g, \bar{Y}^2_g) \rbrace$ as described above where $j$ now indexes experimental groups.
\item Compute $\hat{\beta}_{q}$ for a variety of thresholds $q$ using $\lbrace (\bar{X}^1_g, \bar{Y}^1_g) \rbrace$.
\item Compute treatment level predictions of $Y$ using fold $1$ for each level of $q$: $\hat{Y}^q_g = \bar{X}^1 \hat{\beta}_q$.
\item Choose $q$ which minimizes $\text{IVCV}(q) = \sum_{j} (\bar{Y}^2_g - \hat{Y}^q_g)^2.$
\end{enumerate}

The intuition behind IVCV is similar to the main idea behind IV in general. Recall that our objective is to use variation in $X$ that is not caused by $U$. The IVCV algorithm uses the $X$ value from fold $1$ and compares the prediction to the $Y$ value in fold $2$ because fold $1$ and fold $2$ share a $Z$ but differ in $U$ (since $U$ is independent across units but $Z$ is the same within group). This intuition has been exploited in split-sample based estimators \citep{angrist1995split, imbens1999jackknife,hansen2014instrumental}.

\begin{figure}[t]
	\begin{center}
		\includegraphics[scale=.4]{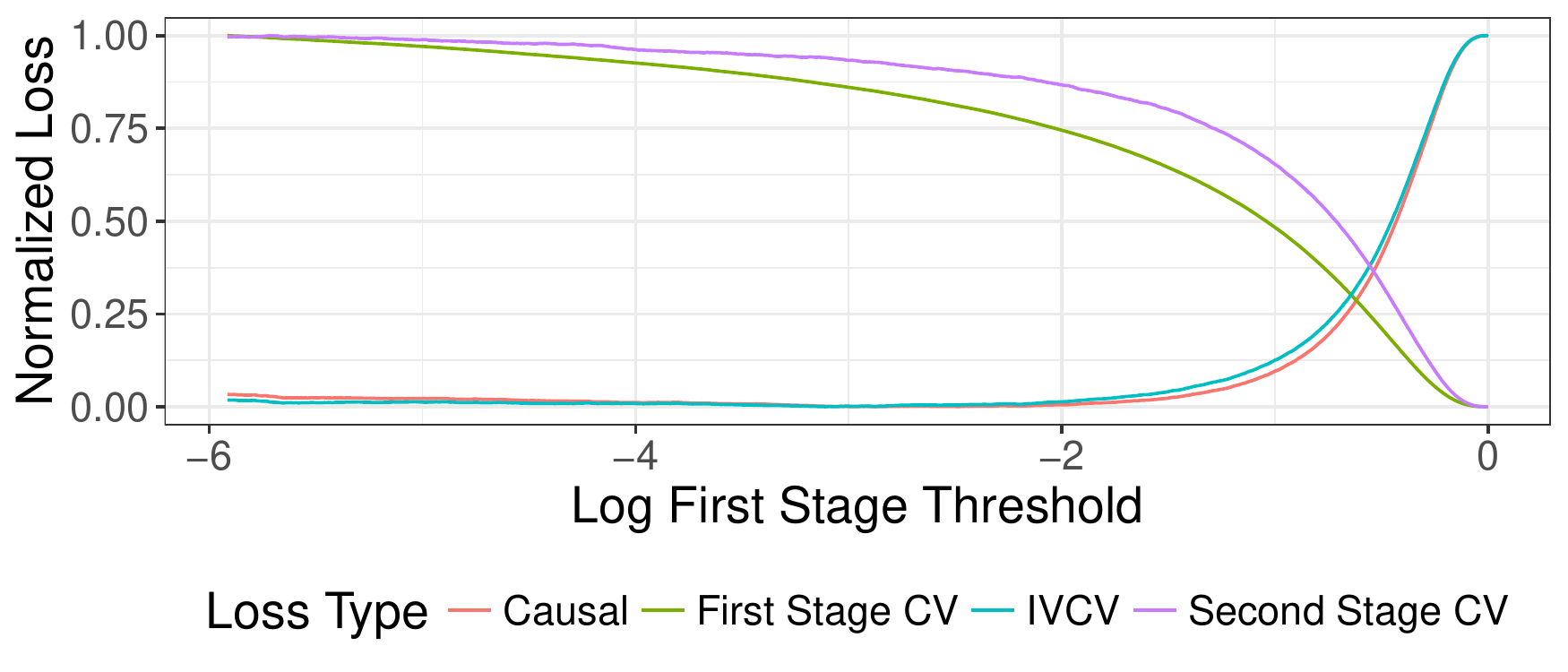}
	\end{center}
	\caption{Comparison of stagewise vs. IVCV method. X-axis is the strength of regularization (lower $p$-value implies stronger regularization). Optimizing for stagewise loss would imply using almost no regularization whereas optimizing for IVCV loss implies strong regularization. Causal loss coincides much more with IVCV loss than stagewise loss.}
	\label{fig:ivcv_vs_stagewise}
\end{figure}

We can demonstrate the importance of using the full causal loss by comparing the IVCV procedure to other two candidates. The first is simply applying naive CV in the second stage (i.e., splitting each group into 2, training a model on fold 1 and computing the CV loss naively as $\| Y_2 - X_2 \hat{\beta}_q \|^2$). The second is stagewise, in which the regularization parameter is chosen to minimize MSE in the first stage, and then the second stage is fit conditional on the selected model \citep[as in][]{belloni2012sparse,hartford2016deepiv}. We compare these approaches in a simple linear model with scalar $X$, such that $\bar{Y} = \bar{X} + \bar{U} \gamma$ and  $\bar{X} = \bar{Z} + \bar{U}$) with $\bar{Z} = \E[X \mid Z]$ distributed $t$ with 3 degrees of freedom and scale $.4$, $\gamma=10,$ $\nper = 100$ and $K=2500$. 

Figure \ref{fig:ivcv_vs_stagewise} shows naive (second stage) CV loss $(Y_2 - X_2 \hat{\beta})^2$, first stage CV loss $(X - \hat{X})^2$, true causal loss $(\beta - \hat{\beta})^2$, and IVCV loss as a function of the first stage regularization parameter averaged over $500$ simulations of the model above. We see that both the first stage loss curve and the naive CV loss curve look very different from the causal loss curve. However, the IVCV loss curve matches almost exactly. Thus, either stage error naively yields a very different objective function from minimizing the causal error. In particular, we see that making the bias--variance tradeoffs for the first stage need not coincide with an desirable bias-variance tradeoff for causal inference.

The $l_0$-regularized IV estimator only requires summary statistics per experimental group that are already routinely computed in the course of running A/B tests. However, IVCV as specified above requires uses raw data. In the Supplemental Material we show that IVCV can also be implemented using only summary statistics. This is because the distribution of two normal random variables which sum to another normal random variable has a closed form from which it is easy to sample. Thus, the full procedure is implementable using a highly compressed form of the original data.

\section{Evaluation}
We now evaluate these procedures empirically. True causal effects in real data are generally unobservable, so comparisons of methods usually lack a gold standard.\footnote{Examples of the kinds of evaluations usually done include: comparing different observational procedures to what is estimated by an experiment or comparing different procedures and showing that one yields estimates which are more `reasonable.'} On the other hand, simulations allow us to know the true causal effects, but can lack realism. We strike a middle ground by using simulations where we set the causal effects ourselves but other joint distributions are determined by a collection of real randomized experiments. These simulations use a model given by $\bar{X} = \bar{Z} + \bar{U} \text{ and } \bar{Y} = X \beta + \bar{U} \gamma.$ Thus, in this case all the variance in $X$ that is not driven by our instruments is confounding variance.

\subsection{Data}
The multivariate case is made difficult and interesting when $U$ has a non-diagonal covariance matrix and $\bar{Z}$ has some unknown underlying distribution, so we generate these distributions from real data derived from 798 randomly assigned test groups from a sample of Facebook A/B tests.\footnote{Note that we use the collection of A/B tests only to generate a distribution for our first stage (i.e., $\E[X \mid Z]$). In the Supplement Material we also consider the IVCV procedure in several completely synthetic data sets. The synthetic data allows us to elucidate the important assumptions for our procedure to work while the main evaluation shows that these assumptions are indeed satisfied in real world conditions.} We define our endogenous, causal $X$s as 7 key performance indicators (i.e. intermediate outcomes examined by decision-makers and analysts); we standardize these to have mean 0 and variance 1. As the distribution of $U$ we use the estimated covariance matrix among these outcomes in observational data. Third, we take the experiment-level empirical means of the $X$s as the true $\bar{Z}$, to which we add the confounding noise according to the distribution of $U$.

\begin{figure}[t]
\begin{center}
	\includegraphics[width=.4\columnwidth]{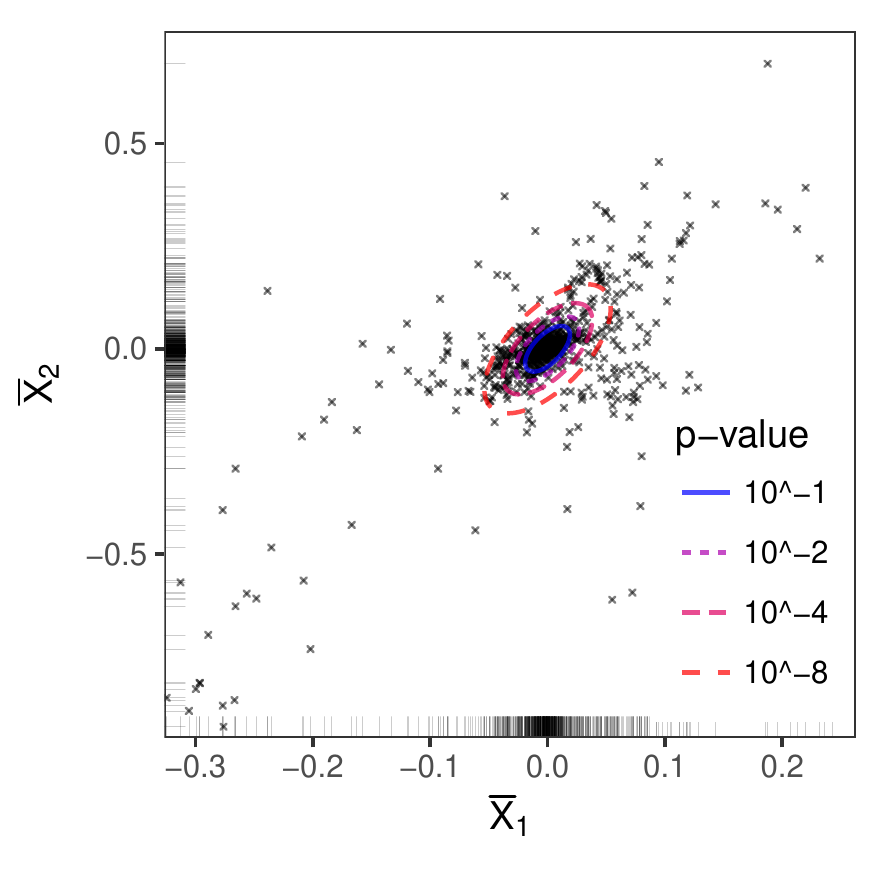}
	\includegraphics[width=.395\columnwidth]{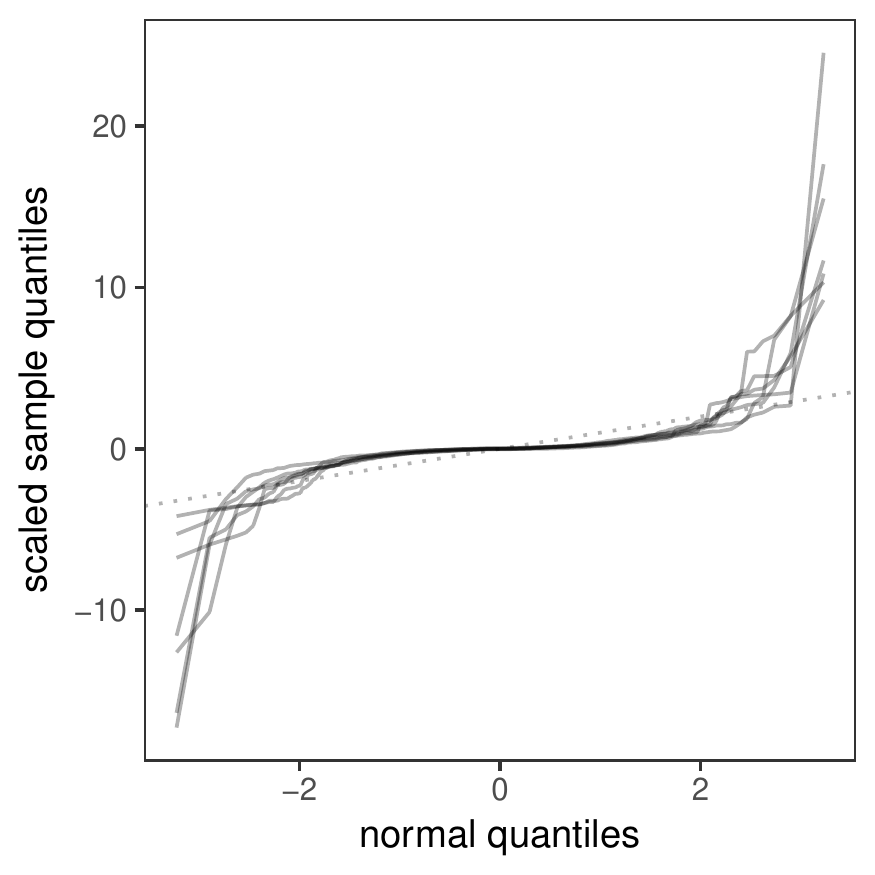}
	\end{center}
	\caption{
	\textbf{A)} Two dimensions of the multivariate means for sampled test groups ($\bar{X}_g$).
	\textbf{B)} QQ-plots for the dimensions of the sampled test groups ($\bar{X}_g$). The marginal distributions are notably non-normal.
	}
	\label{fig:data}
\end{figure}

We show a projection of these $\bar{Z}$ onto $2$ of the $X$ dimensions in Figure \ref{fig:data}(A). We see that the A/B tests appear to have correlated effects but do span both dimensions independently, many groups are retained even with strong first stage regularization, and the distribution has much more pronounced extremes than would be expected under a Gaussian model. Figure \ref{fig:data}(B) compares the observed and Gaussian quantiles, illustrating that all dimensions are notably non-normal (Shapiro--Wilk tests of normality, all $p$s $ < 10^{-39})$.

We set $\beta$ as the vector of ones and $\gamma$ as a diagonal matrix with alternating elements $1$ and $-1$, so that there is both positive and negative confounding. For each simulated data set, we compute the causal mean squared error for $\beta$; that is, the expected risk from intervening on one of the causal variables at random. If $\hat{\beta}$ is our estimated $\beta$ vector then this is $\| \hat{\beta} - \beta \|^2.$

\begin{figure}[t]
\begin{center}
	\includegraphics[width=.6\columnwidth]{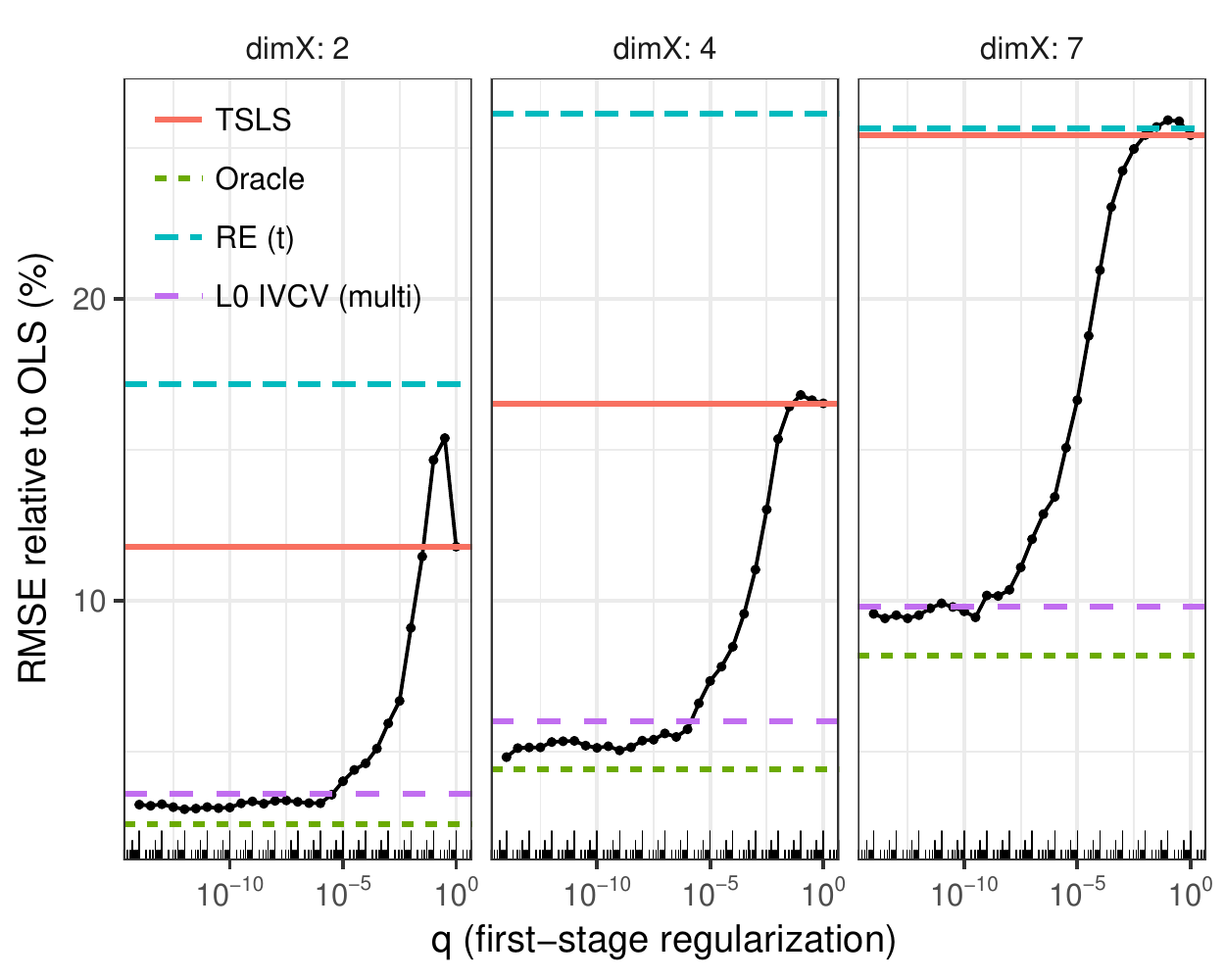}
	\includegraphics[width=.34\columnwidth]{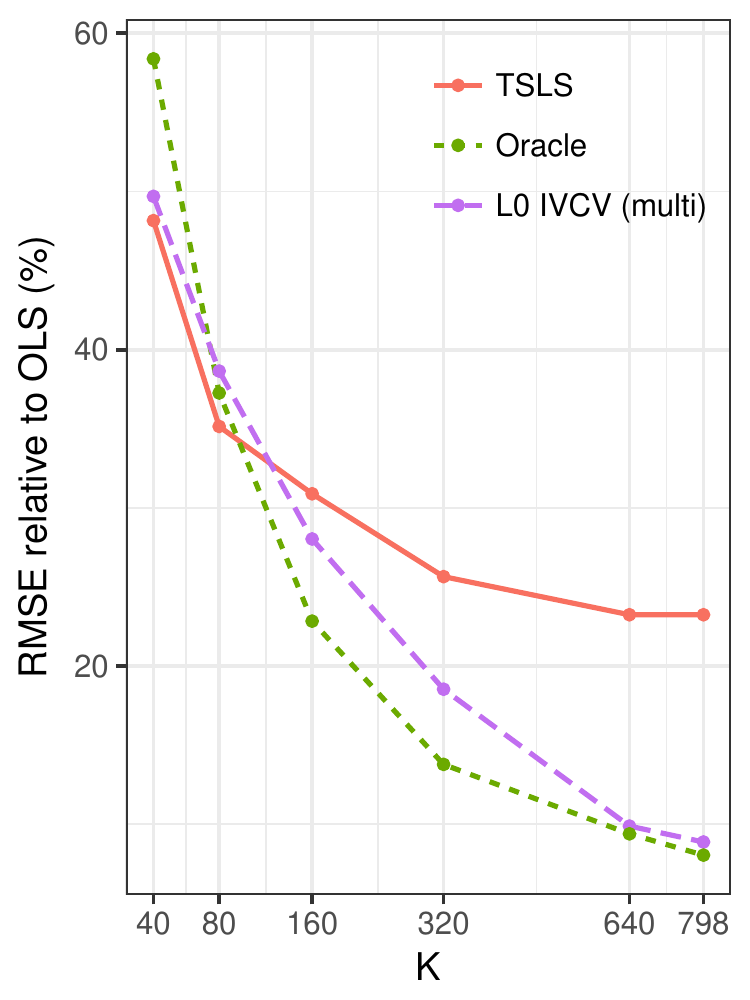}
	\end{center}
	\caption{
	\textbf{A)} Causal error (relative to a naive observational estimator) for the full $l_0$-regularization path (solid black), TSLS (solid red), IVCV selected parameters (dashed purple) and Bayesian random effects model (dashed teal). IVCV outperforms all other estimation techniques.
	\textbf{B)} Error in estimating causal effects for varying numbers of test groups $K$. IVCV is useful even with a relatively small meta-analysis, while TSLS exhibits asymptotic bias. With a very small number of test groups, the Oracle can actually underperform TSLS because of near collinearity.
	}
	\label{fig:main_expt}
\end{figure}

\subsection{Results}
In addition to the $l_0$-regularized IV method and TSLS, we examine a Bayesian random effects model, as in \citet{chamberlain2004random} but with a $t$, rather than Gaussian, distribution for the instruments. Let $\bar{Z} \sim t(d)$ with the prior for $d \sim \text{Gamma}(2, .2)$ (a standard prior in the literature). We also give the model the true covariance matrix for $\bar{U}$. To fit the model we use Stan~\cite{carpenter2016stan}. We compare the Bayesian random effects model and our regularized IV model to the infeasible Oracle estimator where the estimate of the first stage $\E [\bar{X} \mid \bar{Z}]$ is known with certainty.

Figure \ref{fig:main_expt}(A) shows the results for various dimensions of $X$ for 1,000 simulations. Because of the high level of confounding in the observational data, the observational (OLS) estimates of the causal effect are highly biased, such that even the standard TSLS decreases our causal MSE by over $70\%.$ 

We see that the $l_0$-regularization path (black line) reduces error compared with TSLS and, with high regularization, approaches the Oracle estimator. Furthermore, feasible selection of this hyperparameter using IVCV leads to near optimal performance (purple line). The Bayesian random effects model can reduce bias, but substantially increases variance and thus MSE. 

We also look at how large the collection of experimental groups needs to be to see advantages of a regularized estimator relative to a TSLS procedure. We repeat the TSLS, Oracle, and $l_0$-regularization with IVCV analyses in 100 simulations with smaller $K$ (Figure \ref{fig:main_expt}(B)) for the case of the $7$ dimensional $X$. 
Intuitively, what is important is the relative size of the tails of the distribution of the latent treatment effects $\bar{Z}$. As the tails get fatter, fewer experiments are required to get draws from the more extreme components of the mixture. We see that in this realistic case where $\bar{Z}$ is determined using a sampled set of Facebook A/B tests, feasible selection of the $l_0$-regularization hyperparameter using IVCV outperforms TSLS substantially for many values of $K$. Thus, meta-analyses of even relatively small collections of experiments can be improved by the first-stage $l_0$ regularization.

\section{Conclusion}
Most analyses of randomized experiments, whether in academia, business, or public policy tends to look at each trial in isolation. When meta-analyses of experiments are conducted, these usually either pool data about multiple instances of the same intervention or to find heterogeneity in the effects of interventions across settings or methods \citep[e.g.,][]{hemkens2016agreement}. We instead propose combining many experiments can help us learn richer causal relationships that are not identified by any single experiment. IV models give a way of doing this pooling. We have shown that in such situations using easily-implemented $l_0$ regularization reduce bias and total error in estimating causal effects, and thus produce better predictions about interventions, than using standard TSLS methods. 

We expand on the literature which uses multi-condition experiments as instruments \citep{eckles2016estimating, goldman2014experiments}. Such analyses feature a smaller number of experimental groups and a single causal variable. Our work is also related to research on IV estimation with weak instruments \citep{stock2012survey, staiger1997instrumental, stock2005testing}. In addition, we also contribute to existing research on regularized IV estimation \citep{belloni2012sparse, hansen2014instrumental, chamberlain2004random}.
Our application domain motivates introducing a group-$l_0$ regularization and a feasible and data efficient cross-validation procedure, while previous techniques have used naive stagewise cross-validation.

The present work is part of a growing literature on machine learning techniques and causality \citep{bottou2014machine}, much of which has focused on learning causal graphs \citep{pearl2009causality}, observational causal inference \citep{shalit2016bounding}, heterogeneous treatment effects \citep{grimmer2014estimating, athey2016recursive, peysakhovich2016combining}, or contextual bandit problems \citep{agarwal2014taming, dudik2014doubly,swaminathan2015counterfactual}, but only more recently on instrumental variables methods \citep{hartford2016deepiv,peters2016causal}.

%Recently, active learning in the form of bandit optimization \citep{gittins2011multi, li2010contextual} and reinforcement learning \citep{sutton1998reinforcement} have become quite popular in the AI community. These approaches can be used in many of the same contexts as the IV analysis we have discussed here, and so may appear to be competitors. However, we note there are many important differences the largest of which is that RL and bandit approaches try to perform policy optimization rather than learning of a causal graph. For this reason, often \textit{why} RL/bandit estimated policies work can be hard to understand. This is in contrast to explicit causal models (e.g., the linear model described above) which can be stated explicitly and in terms that are more natural to human decision-makers \citep{jung2017simple}. On the other hand, bandit/RL approaches have advantages in that they are explicitly online whereas many causal inference procedures (including the one we have described here) are `batch' procedures that assume that data collection is a passive enterprise, separate from analysis. There is growing interest in combining these approaches \citep{lattimore2016causal} and we think that future work would benefit greatly from this fusion of thought.

\bibliography{BFIV_nips_submit.bbl}
%\bibliography{bfiv_paper.bib}
\bibliographystyle{icml2016}

\section{Supplemental Material}
\subsection{Confounding in a Linear Model}

\begin{figure}[h!]
	\begin{center}
		\includegraphics[scale=.20]{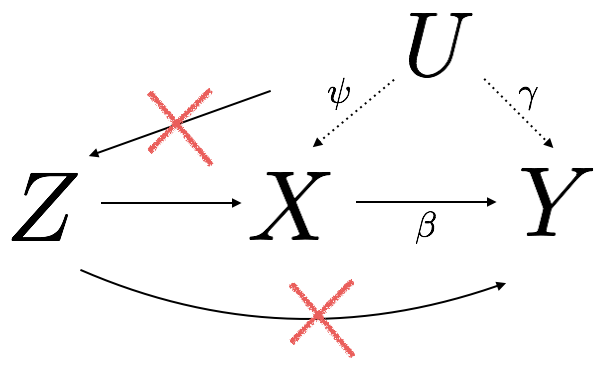}
	\end{center}
	\caption{DAG representing our structural equations, in which the relationship between $X$ and $Y$ is confounded by $U$, and including the instrumental variable $Z$. Crosses represent causal relationships that are ruled out by the IV assumptions.}
	\label{fig:iv_setup}
\end{figure}

Consider the linear structural equation pair from the main text:
$$X = U\psi + \epsilon_{X}$$
$$Y = X\beta + U\gamma + \epsilon_{Y}$$
where these variables have mean 0 and finite variances $\sigma^2_U, \sigma^2_{\epsilon_{X}}$ and $\sigma^2_{\epsilon_{X}}$.

Suppose that we only observe $(X,Y)$ where both are scalar. Since the underlying model is linear, we can try to estimate it using a linear regression. However, not including the confounder $U$ in the regression yields the estimator: 
\begin{equation}
\label{eq:ols}
\hat{\beta}_\text{obs} = (X' X)^{-1} (X' Y)
\end{equation}
When all variables are scalar algebra yields
$$
\E [\hat{\beta}_\text{obs}] = \beta + \gamma \frac{\Cov(X,U)}{\Var(X)}.
$$ 

\subsection{Derivation of the Group IV Bias}
Let us use the convention from the main text and denote by $\bar{A}$ the group level mean of variable $A.$ This means we get 

$$\bar{X} = \bar{Z} + \bar{U} \psi + \epsilon_{\bar{X}}$$ $\bar{Y} = \bar{X} \beta + \bar{U} \gamma + \epsilon_{\bar{Y}}$ 

Since the TSLS estimator in this case is a regression of $\bar{X}$ on $\bar{Y}$ we can use the equation derived above for the scalar case to rewrite $$\E [\beta_{TSLS}] = \beta + \gamma \frac{\Cov(\bar{X}, \bar{U})}{\Var(\bar{X})}.$$

\subsection{IVCV With Only Summary Statistics}
The $l_0$-regularized IV estimator only requires the kinds of summary statistics per experimental group that are already recorded in the course of running A/B tests, which has practical and computational utility. However, the cross-validation procedure above requires the use of raw data. We now turn to the following question: if the raw data is unavailable, but summary statistics are, can we use these summary statistics to choose a threshold $q$?

Suppose that we have access to summary means $\lbrace (\bar{X}_g, \bar{Y}_g) \rbrace$ for each treatment $j$ and the covariance matrix of $(\bar{X}, \bar{Y})$ conditional on $Z=0$ which we denote by $\tau$. We note that $\tau$ can be estimated very precisely from observational data or, in the case of the experimental meta-analysis just looking at covariances among known control groups. We assume that $\nper$ is large enough such that the distributions of $U$ and $\epsilon$ in groups of size $\frac{\nper}{2}$ are well approximated by the Gaussian $\mathcal{N}(0, \frac{\sigma^2_i}{\frac{\nper}{2}}).$ 

To perform IVCV under these assumptions, we use a result from the literature on Monte Carlo \citep[ch. 8]{owen_mcbook}. If some vector $X$ is distributed multivariate normal $(\mu, \Sigma)$ then any linear combination $T = \theta X$ has a normal distribution. Moreover, conditional on $T=t$ the distribution of $X$ is normal with mean $\mu + \Sigma \Theta' (t - \theta \mu)$ and covariance matrix $\Sigma - \Sigma \Theta' (\Theta \Sigma \Theta')^{-1} \Theta \Sigma.$ 

This means if we know the observational covariance matrix $\tau$ then for every group $g$ we can take the group level averages $(\bar{X}_g, \bar{Y}_g)$ and sample using the equation above to get $\bar{X}^1_g$ and $\bar{X}^2_g$ such that $\bar{X}^1_g + \bar{X}^2_g = 2\bar{X}_g$. Since by the central limit theorem the generating Gaussian model is approximately correct, this procedure simulates the split required by IVCV without having access to the raw data.

This gives us a summary-statistics-based IVCV algorithm:

\textbf{Summary statistics instrumental variables cross-validation algorithm} (sIVCV):
\begin{enumerate} 
\item Start with data comprising of treatment group means $\lbrace (\bar{X}_g, \bar{Y}_g) \rbrace$.
\item Use the covariance matrix to perform Monte Carlo sampling to simulate groups $\lbrace (X^1_i, Y^1_i, Z^1_i) \rbrace$ and $\lbrace (X^2_i, Y^2_i, Z^2_i) \rbrace$.
\item Use the IVCV algorithm to set the hyperparameter using the simulated splits.
\item Estimate $\beta$ using the selected hyperparameters on the full data set.
\end{enumerate}

\subsection{Synthetic IVCV Experiments}
In addition to the real data that we have provided in the main text, we also consider the IVCV procedure in several completely synthetic data sets. This allows us to elucidate the important assumptions for our procedure to work while the main experiment shows that these assumptions are indeed satisfied in real world conditions. 

We consider the same exact model as in the main text except we generate the first stage effects $\bar{Z}$ from a known parametric distribution and let $U$ be normal. First, we consider $X = \bar{Z} + U$ where the treatment effect $\bar{Z}$ is drawn from an independent $t$ distribution with $3$ degrees of freedom. Second, we consider $X = \bar{Z} + U$ where $\bar{Z}$ is drawn from a $t$ distribution with $3$ degrees of freedom with a covariance matrix drawn from an inverse Wishart (a conjugate prior for covariance matrices and a standard way of generating covariance matrices) with $10 \times$ dim$(X) $ degrees of freedom. Note that in former case effects are axis aligned while in the latter case larger values of one dimension can predict more extreme values of $Z$ (and $X$) on another dimension. 

Finally, we consider a model where first we draw a variance $\sigma^2$ from an inverse gamma distribution then we draw $\bar{Z}$ from an independent normal distribution with variance $\sigma^2.$ This means that components are mean-uncorrelated, but that one when component's value is extreme, it is more likely that other components' values are extreme. This is the multivariate analog of our motivating example where some A/B tests are strong explorations of the parameter spaces and others are micro-optimizations at the margin. Note that the marginal distribution for each dimension is, just like in the first example, a $t$ distribution with $3$ degrees of freedom (since the $t$ can be written as a mixture of normals drawn from the inverse gamma). 

\begin{figure}[h!]
	\begin{center}
		\includegraphics[width=.45\columnwidth]{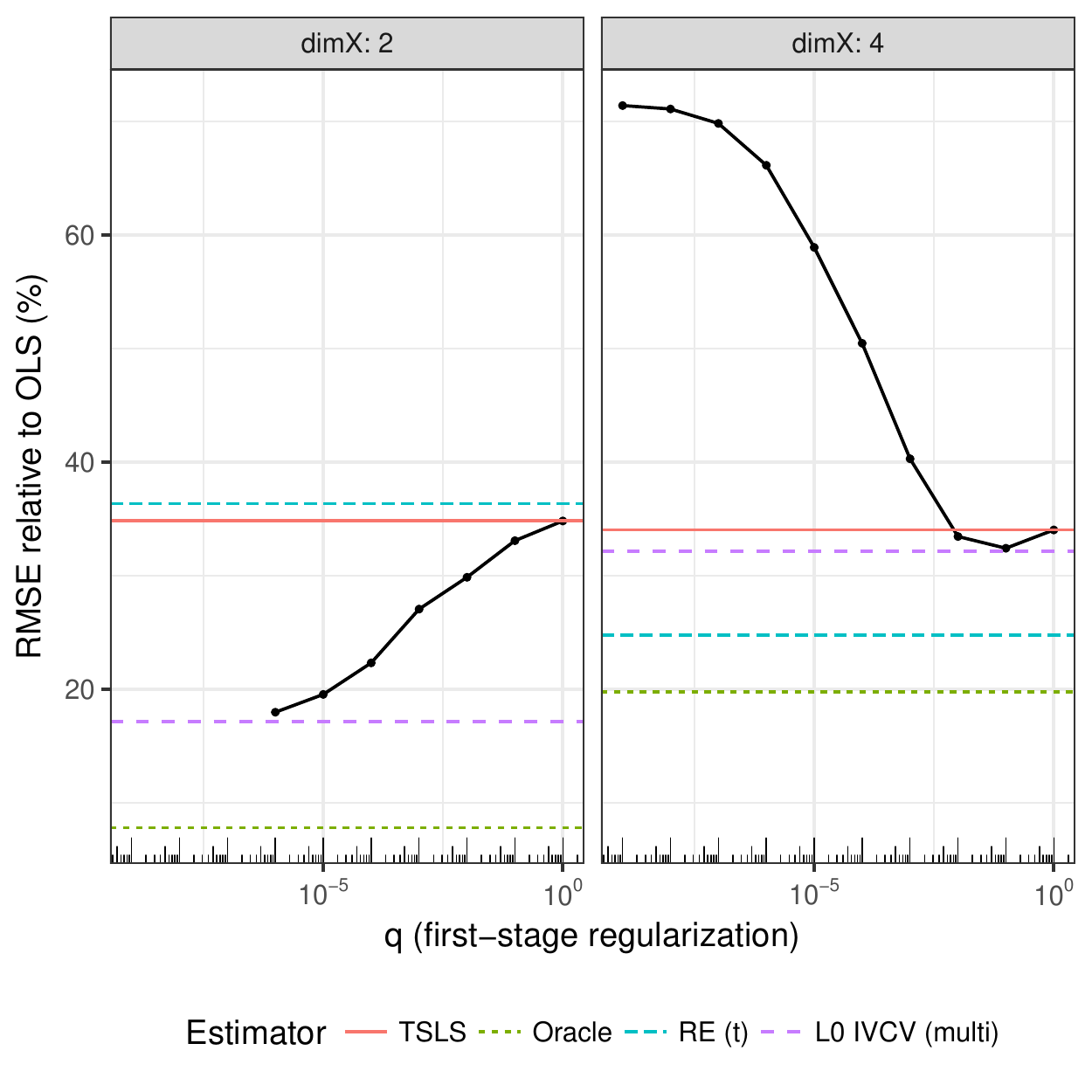}
		\includegraphics[width=.45\columnwidth]{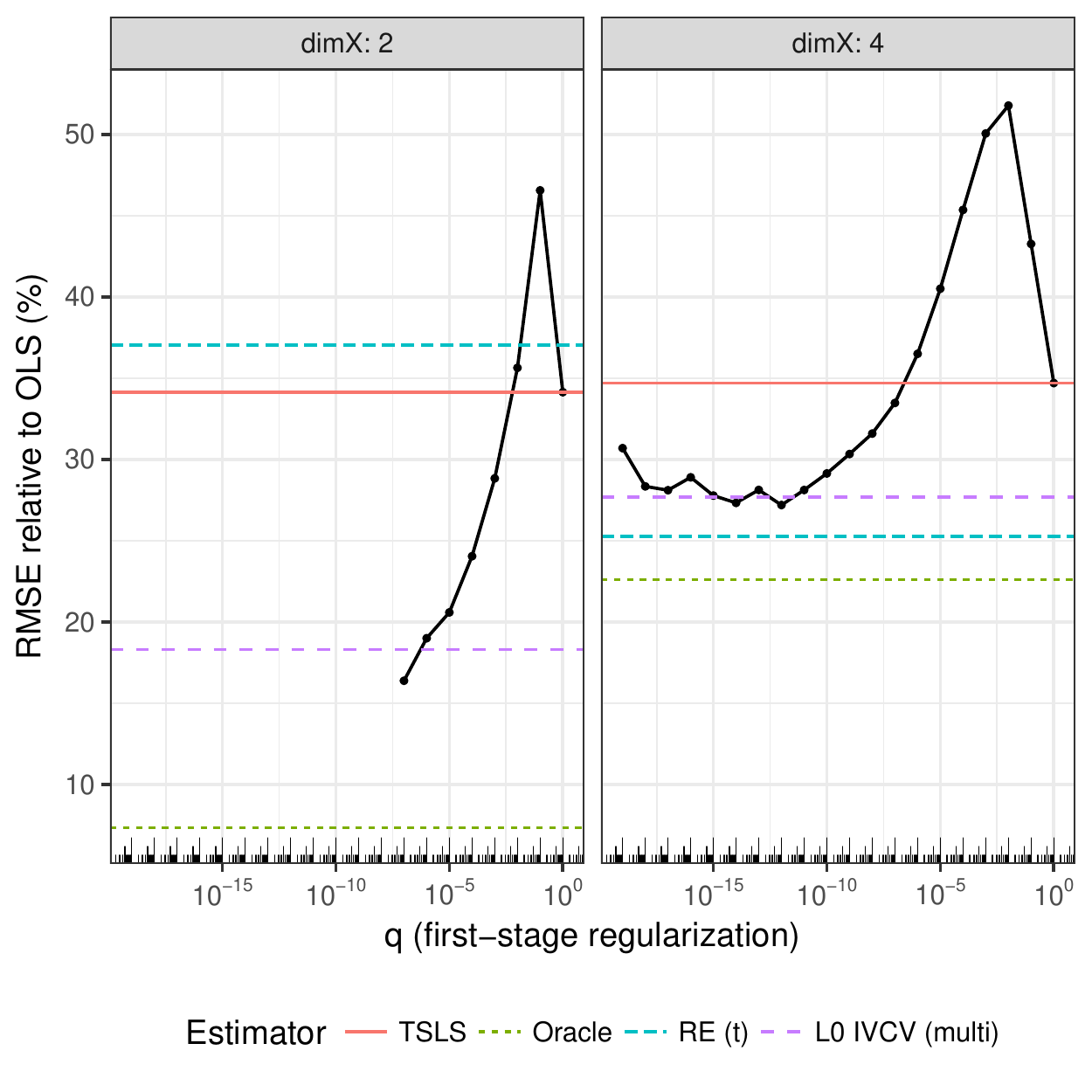}
		\includegraphics[width=.45\columnwidth]{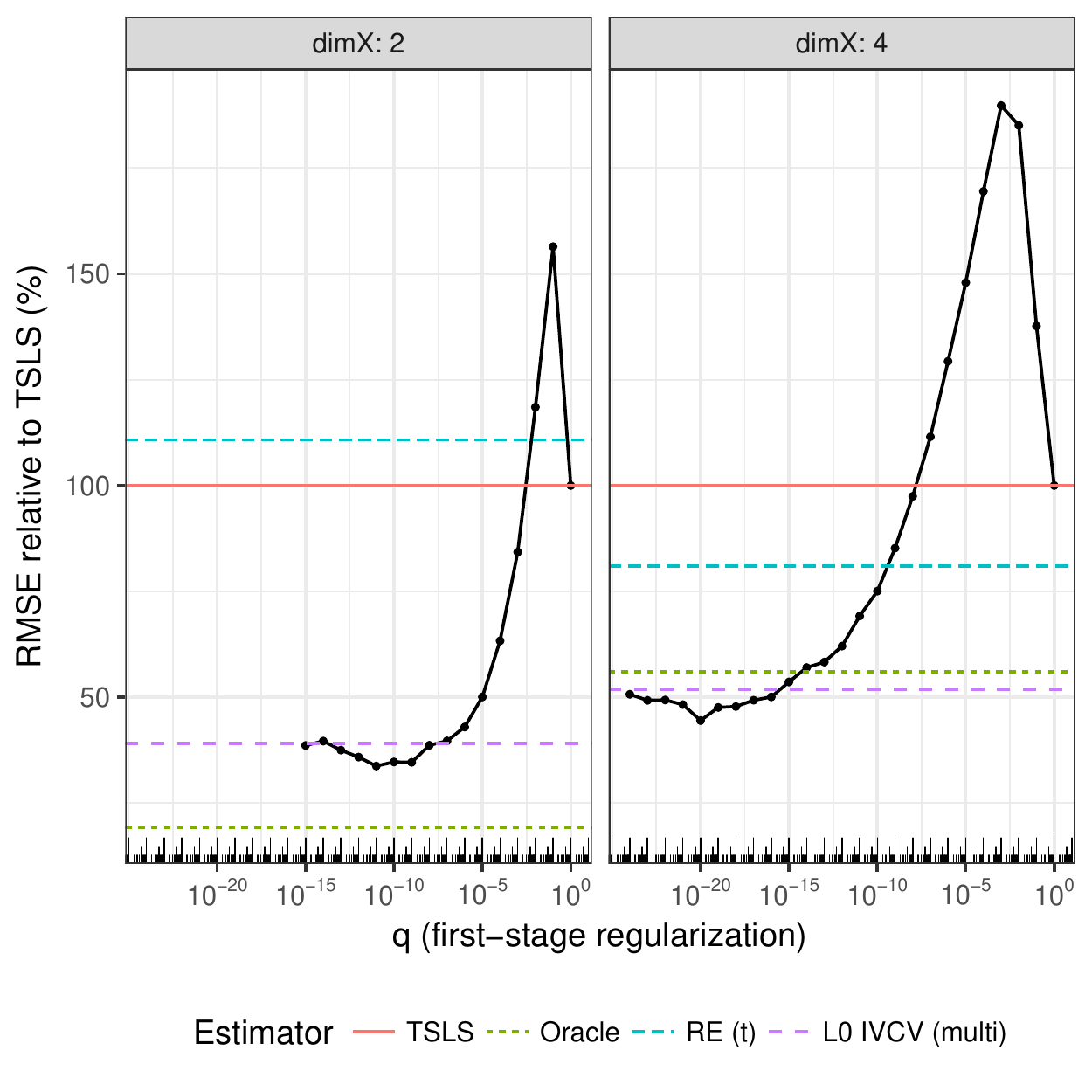}
	\end{center}
	\caption{Performance of various IV estimation techniques under various first stage data generating assumptions (left = independent $t$, right = Wishart $t$, bottom = correlated variances). We see that when the $Z$ induced components of $X$ are independent even for moderate dimensionality that the $l_0$ regularization performs less well. However, as soon as there is any correlation the IVCV procedure performs much better than TSLS and can both under or over-perform the Bayesian random effects model. In the main text we see that in a real distribution the IVCV does indeed beat the Bayesian model.}
	\label{fig:tsims}
\end{figure}

Figure \ref{fig:tsims} shows key main text figure replicated using the data generating processes above (left = independent $t$, middle = Wishart $t$, right = correlated variances). We restrict to $\text{dim}(X) \in \lbrace 2, 4 \rbrace$ because it is sufficient to illustrate our main points. We see that in the independent $t$ case the IVCV procedure (and indeed our multivariate $l_0$ regularization) can underperform the Bayesian random effects model fail to substantially improve on TSLS. This happens because in the independent $t$ case there is a high probability that a single dimension is extreme enough to pass the regularization threshold and thus even strong regularization does not necessarily remove bias. On the other hand, when outcomes are correlated (or their variances are) we see that multivariate IVCV performs well because being extreme in one $X$ component predicts having extreme outcomes in other components. This leads to an interesting question of whether there is a more efficient regularization design. 

\end{document}